\documentclass[10pt,twocolumn,letterpaper]{article}

\usepackage[pagenumbers]{cvpr}
\usepackage{colortbl}
\usepackage{multirow}
\definecolor{TableHeaderGray}{HTML}{F2F2F2}
\definecolor{TableAccentBlue}{HTML}{E8ECFF}
\definecolor{cvprblue}{rgb}{0.21,0.49,0.74}
\usepackage[breaklinks,colorlinks,allcolors=cvprblue]{hyperref}

\def\confName{CVPR}
\def\confYear{2027}

\title{V-FIND: Revealing the Intrinsic Forgery Knowledge Encoded \\ in Video Forgery Detectors}

\author{
Shichao Kan\textsuperscript{1}, Chengpeng Hong\textsuperscript{1}, Jingtong Dou\textsuperscript{2}, Chuancheng Shi\textsuperscript{2},\\
Yuhan Liu\textsuperscript{1}, Linrui Xu\textsuperscript{1}, Yixiong Liang\textsuperscript{1}, Yigang Cen\textsuperscript{1},\\
Yanpeng Sun\textsuperscript{3}, Fei Shen\textsuperscript{3*}, Tat-Seng Chua\textsuperscript{3}\\[0.4em]
{\small \textsuperscript{1}Central South University \quad
\textsuperscript{2}The University of Sydney \quad
\textsuperscript{3}NExT++ Research Centre, National University of Singapore}\\
{\small \textsuperscript{*}Corresponding author}
}

\begin{document}
\maketitle

\begin{abstract}
As generated videos become increasingly realistic, reliable video forgery detection is increasingly important. Existing studies typically optimize and use video forgery detectors as black boxes, while the latent forgery-discriminative knowledge inside them remains largely unexplored. Instead of continuing to rely on resource-intensive full-model retraining to steadily improve detection performance, we ask whether video forgery detection can also be achieved by uncovering and activating sparse forensic knowledge within the detector. We find that a sparse set of functionally specialized neurons encodes more forgery-discriminative knowledge, enabling a compact and effective forensic readout. Based on this insight, we propose a video forgery-intrinsic neuron discovery (V-FIND) framework for video forgery detection. V-FIND first localizes critical layers that exhibit discrepancies between real and forged videos, and then identifies latent anchor neurons that consistently carry forgery-discriminative signals, forming a compact forensic subspace. With the original backbone frozen and only a lightweight linear classifier trained, this subspace still delivers strong detection performance across multiple external benchmarks for generated videos. Further neuron intervention experiments provide direct evidence for the functional specificity of the discovered neurons. Overall, these results suggest that a compact subset encodes more forgery-discriminative knowledge that is extractable and reusable, offering a new perspective on understanding and exploiting the intrinsic forensic capability of video forgery detectors.
\end{abstract}

\section{Introduction}
Artificial intelligence-generated content (AIGC)~\cite{11304732} has advanced rapidly in recent years, driving the creation and deployment of high-quality synthetic media. This evolution is particularly pronounced in the field of video generation. Driven by breakthrough technologies such as latent and video diffusion models~\citep{ho2022video}, text-to-video generation pipelines, diffusion transformers (DiTs)~\citep{peebles2023scalable}, and large-scale world simulators~\citep{brooks2024videoworld}, contemporary commercial engines are now capable of synthesizing visually flawless videos.
As synthetic videos become increasingly photorealistic, they pose more immediate threats to media authenticity, information security, and public trust. Reliable and effective video forgery detection has therefore become an important and urgent problem.

\begin{figure}[t]
\centering
\includegraphics[width=0.98\linewidth,trim=94.4pt 45.2pt 274pt 59.6pt,clip]{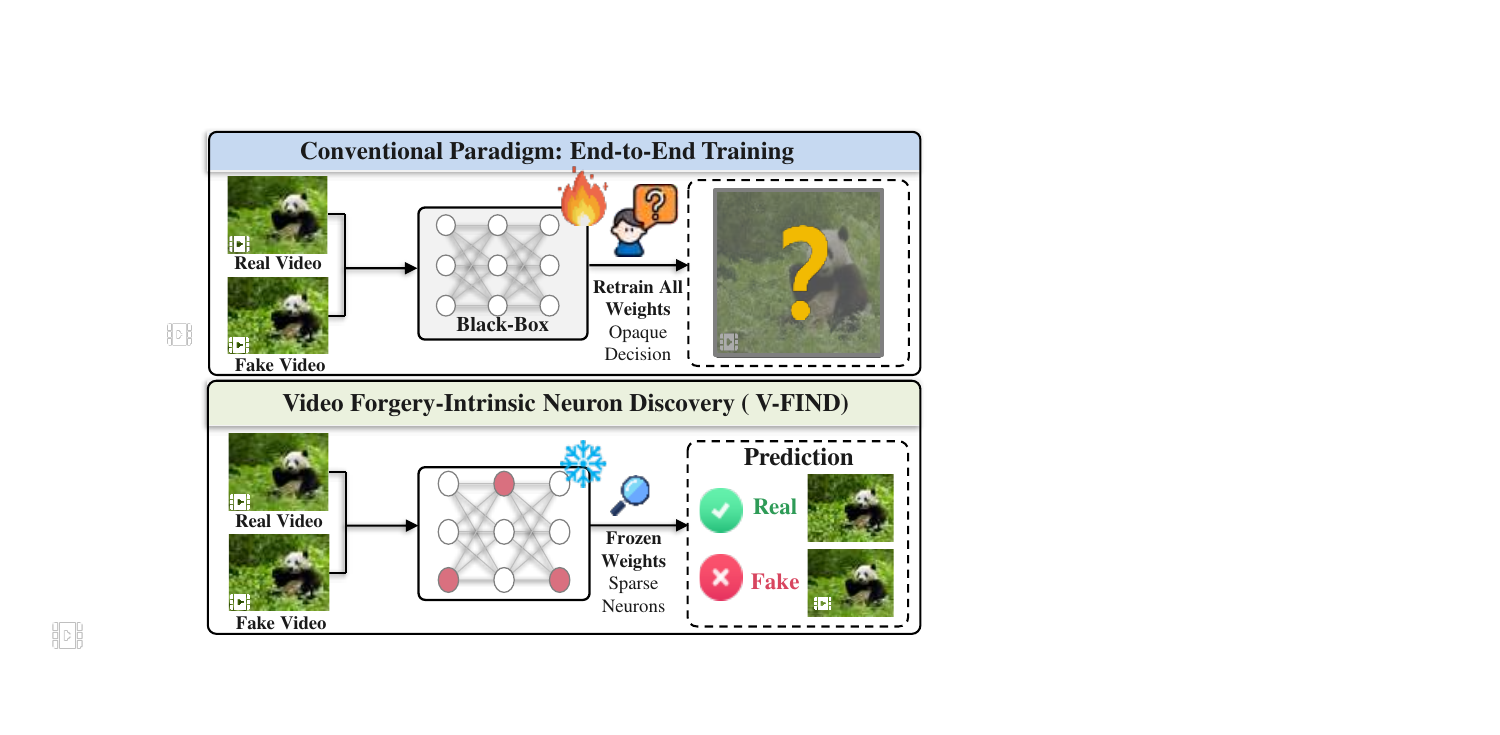}
\vspace{-0.3cm}
\caption{\textbf{Comparison of Conventional End-to-End Training and Our V-FIND Paradigm.} Conventional paradigm retrains all backbone weights as an opaque black box. Conversely, V-FIND uncovers sparse intrinsic forgery neurons within a frozen backbone for parameter-efficient detection.}
\label{fig:motivation}
\vspace{-0.5cm}
\end{figure}

In response to these escalating threats, the defense landscape has dynamically evolved over the years from deepfake detectors~\citep{qian2020thinking,yan2023deepfakebench,xu2023tall} focused on facial manipulations to more recent detectors for fully AI-generated videos~\citep{ma2024decof,ni2025genvidbench,ji2024distinguish,song2024multimodal,chen2024demamba,corvi2025seeing,kundu2025universal,li2026vina,li2026preserving}. While this steady progression has pushed benchmark performance to new heights, it has largely relied on a homogeneous recipe of scaling up visual backbones, expanding training corpora, and performing resource-intensive end-to-end optimization. Consequently, existing studies typically optimize and use video forgery detectors as indivisible black boxes, while the latent forgery-discriminative knowledge encoded within them has yet to be systematically uncovered and exploited. This gap prompts us to rethink video forgery detection from an intrinsic-knowledge perspective in this paper. Instead of continuing to rely on full-model retraining to steadily improve detection performance, we ask whether video forgery detection can also be achieved by uncovering and activating sparse forensic knowledge within the detector. Figure~\ref{fig:motivation} contrasts this perspective with the conventional end-to-end detector paradigm.

Under this rethinking, we shift the focus from optimizing the detector as an indivisible whole to examining how forgery-discriminative knowledge is organized within its internal representation space. Our central view is that a sparse set of functionally specialized neurons encodes more forgery-discriminative knowledge. If these neurons consistently carry forgery-discriminative signals, they should be explicitly localizable and could be organized into a compact forensic subspace. This possibility suggests a complementary route to video forgery detection: rather than continuing to update the full detector, we seek to uncover and activate its sparse forensic knowledge while keeping the original backbone frozen. If the resulting subspace can still deliver strong detection performance with only a lightweight linear classifier trained, it would indicate that a compact subset of neurons encodes more forgery-discriminative knowledge in an extractable and reusable form. This rethinking therefore shifts the objective from solely optimizing a black-box detector toward understanding and exploiting its intrinsic forensic capability.

To this end, we propose V-FIND, a video forgery-intrinsic neuron discovery framework that uncovers forgery-discriminative knowledge through a coarse-to-fine procedure. V-FIND first localizes critical layers that exhibit pronounced discrepancies between real and forged videos, and then identifies latent anchor neurons (LANs) that consistently carry forgery-discriminative signals, organizing them into a compact forensic subspace. With the original backbone frozen and only a lightweight linear classifier trained, this subspace still delivers strong detection performance across multiple external benchmarks for generated videos. Further neuron intervention experiments provide evidence for the functional specificity of the discovered neurons. Overall, these results suggest that video forgery detectors contain sparse, extractable, and reusable forgery-discriminative knowledge, offering a new perspective on understanding and exploiting their intrinsic forensic capability. Our contributions are threefold:
\begin{itemize}
\item We show that a sparse set of functionally specialized neurons encodes more forgery-discriminative knowledge, motivating an intrinsic-knowledge perspective for video forgery detection.

\item We propose V-FIND, a video forgery-intrinsic neuron discovery framework that localizes critical layers and identifies latent anchor neurons, organizing them into a compact forensic subspace in a coarse-to-fine manner.

\item We show that the forensic subspace achieves strong detection performance with a frozen backbone and a lightweight linear classifier, and neuron intervention verifies the functional specificity of identified neurons.
\end{itemize}

\section{Related Work}
\noindent\textbf{AI-Generated Video Detection.} AI-generated video detection has recently expanded from traditional deepfake detection~\cite{yan2023deepfakebench,xu2023tall} to the detection of fully synthetic videos~\citep{ma2024decof,kundu2025universal,ni2025genvidbench}. Existing methods mainly follow two technical routes. One line focuses on single-frame spatial artifacts~\citep{ji2024distinguish,song2024multimodal,li2026vina}, detecting synthetic content through anomalous textures, local visual cues, or multimodal representations extracted from individual frames. 
Another line emphasizes multi-frame temporal or spatiotemporal inconsistencies~\citep{chen2024demamba,corvi2025seeing,li2026preserving}, improving robustness across generators through stronger video backbones, video-level representation learning, temporal modeling, forensic-oriented augmentation, or native-scale artifact preservation. Despite these different designs, both lines mainly treat the detector as a black-box mapping from input videos to binary labels, focusing on how to build a stronger detector rather than on how forgery-discriminative knowledge is organized across layers and neurons inside the detector.

\noindent\textbf{Neuron Interpretability.} Neuron interpretability studies suggest that model capabilities are often not uniformly distributed across the full representation space, but can be localized to sparse, functionally specialized neurons, concepts, circuits, or subspaces~\citep{bau2017network,kim2018tcav,olah2017feature,alain2016linear,olah2020circuits,bau2020understanding,schwettmann2023multimodal,huang2025neurons,tang2024languagespecific}. Recent work further uses domain-specific neuron discovery and sparse feature decomposition~\citep{huo2024mmneuron,zaigrajew2025hsae,huang2025tide,lim2025patchsae} to reveal specialized internal units. Complementary lines on concept ablation, neuron-level intervention, and circuit discovery~\citep{kumari2023ablating,he2025singleneuron,kwon2025gcc} indicate that manipulating a small subset of internal units can amplify, suppress, or erase specific behaviors, while neuron importance scoring~\citep{xie2021importance} provides practical tools for ranking and selecting task-relevant units. The convergence of these perspectives lays the groundwork for our V-FIND framework, marking a paradigm shift from acquiring external skills through fine-tuning to explicitly awakening intrinsic forgery-discriminative knowledge already encoded within internal representations.

\begin{figure*}[t]
\centering
\includegraphics[width=0.98\linewidth,trim=6pt 0.844pt 10.11pt 0pt,clip]{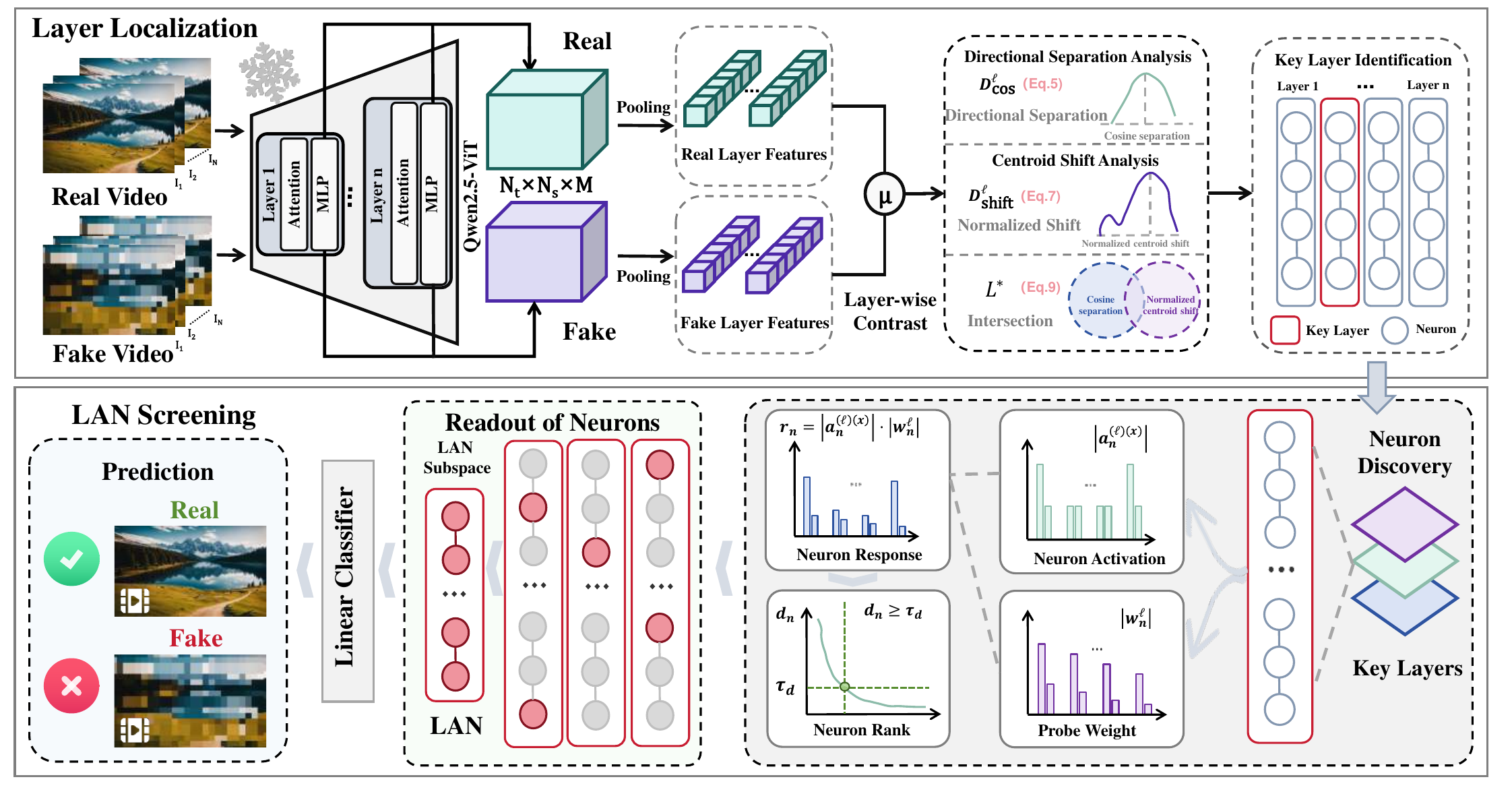}
\vspace{-0.3cm}
\caption{\textbf{Overview of the Proposed V-FIND Framework.} We first localize critical layers by complementary layer-wise discrepancy signals, then identify sparse latent anchor neurons within these layers, and finally train a lightweight detector on the resulting compact forensic subspace.}
\label{fig:method-overview}
\vspace{-0.4cm}
\end{figure*}

\section{Method}
As shown in Figure~\ref{fig:method-overview}, V-FIND follows a coarse-to-fine pipeline. First we localize critical layers by measuring complementary discrepancies between real and forged representations, and then identifies latent anchor neurons (LANs) within these layers using layer-wise probe-based neuron scoring. The selected LAN activations form a compact forensic subspace, which is fed into a lightweight linear classifier while the backbone remains frozen.
Formally, given an input video clip
\begin{equation}
x=\{I_t\}_{t=1}^{T}\in\mathrm{R}^{T\times H\times W\times C},
\end{equation}
where $T$ is the number of sampled frames and $H$, $W$, and $C$ denote the frame height, width, and channels, respectively. The hidden representation at layer  $\ell$ is
\begin{equation}
H^{(\ell)}(x)\in\mathrm{R}^{N_t\times N_s\times M},
\end{equation}
where $N_t$, $N_s$ and $M$ denote the temporal tokens, spatial tokens, and latent dimension, respectively. We obtain the unit-level representation by spatiotemporal average pooling:
\begin{equation}
\mathbf{a}^{(\ell)}(x)=\frac{1}{N_tN_s}\sum_{i=1}^{N_t}\sum_{j=1}^{N_s}H_{i,j}^{(\ell)}(x).
\end{equation}
The scalar $a_k^{(\ell)}(x)$ denotes the response of the $k$-th latent unit at layer $\ell$ for input video $x$. We treat each unit index pair $n=(\ell,k)$ as a candidate latent anchor neuron. This unit-level representation serves as the common internal object for layer localization, neuron discovery, and compact forensic subspace construction.

\begin{figure}[t]
\centering
\includegraphics[width=\linewidth]{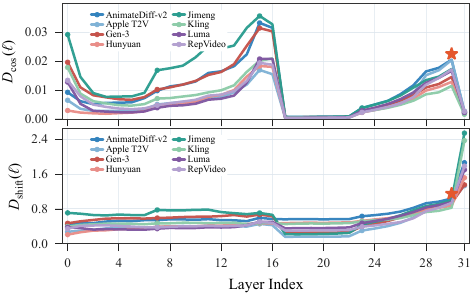}
\vspace{-0.6cm}
\caption{\textbf{Layer Localization.} Complementary layer-wise discrepancy signals across eight fake-source subsets from the detector training distribution. Red stars mark the layer of interest for the subsequent LAN discovery.}
\label{fig:layer-wise-signals}
\vspace{-0.3cm}
\end{figure}

\subsection{Layer Localization}
We first localize layers that exhibit strong real--fake discrepancies, thereby narrowing the search space for subsequent LAN discovery. Specifically, we compute two complementary layer-wise discrepancy signals on the discovery split and retain the layers highlighted by both.
The first signal measures directional separation between real and fake representations. For each layer $\ell$ and class $c\in\{\mathrm{real},\mathrm{fake}\}$, we define the class centroid
\begin{equation}
\mu_c^{(\ell)}=\frac{1}{|\mathcal{D}_c|}\sum_{x\in\mathcal{D}_c}\mathbf{a}^{(\ell)}(x).
\end{equation}
Based on these centroids, we define the directional-separation signal
\begin{equation}
D_{\cos}(\ell)=1-\frac{{\mu_{\mathrm{real}}^{(\ell)}}^\top\mu_{\mathrm{fake}}^{(\ell)}}{\|\mu_{\mathrm{real}}^{(\ell)}\|_2\|\mu_{\mathrm{fake}}^{(\ell)}\|_2},
\end{equation}
which measures how strongly the real and fake centroids diverge in feature orientation. This gives the first candidate layer set
\begin{equation}
\mathcal{L}_{\mathrm{sep}}=\{\ell\mid D_{\cos}(\ell)>\tau_{\mathrm{sep}}\},
\end{equation}
where $\tau_{\mathrm{sep}}$ is determined from the full-layer statistics. This set retains layers whose directional separation is unusually strong relative to the layer distribution of the whole backbone. As shown in Figure~\ref{fig:layer-wise-signals}, the score is concentrated in specific middle and late layers rather than being uniformly distributed across the backbone.

To complement directional separation to fully characterize layer-wise discrepancies, we also measure the normalized centroid shift. Specifically, let $\bar{\sigma}_\ell^2$ denote the mean within-class channel variance at layer $\ell$. It is used to normalize the centroid shift by the within-class variation, and its exact definition is deferred to the appendix. We then define the normalized-shift signal
\begin{equation}
D_{\mathrm{shift}}(\ell)=\frac{\|\mu_{\mathrm{real}}^{(\ell)}-\mu_{\mathrm{fake}}^{(\ell)}\|_2}{\sqrt{M\,\bar{\sigma}_\ell^2}},
\end{equation}
which measures the overall real--fake displacement after controlling for within-class variation. This gives the second candidate set
\begin{equation}
\mathcal{L}_{\mathrm{shift}}=\{\ell\mid D_{\mathrm{shift}}(\ell)>\tau_{\mathrm{shift}}\},
\end{equation}
where $\tau_{\mathrm{shift}}$ is determined from the full-layer statistics. This set highlights layers that still exhibit pronounced class displacement after variance normalization. Figure~\ref{fig:layer-wise-signals} further shows that $D_{\mathrm{shift}}(\ell)$ weakens in the middle stage and then increases steadily in the later layers, suggesting that these layers retain more stable real--fake displacement after accounting for within-class fluctuation.

Finally, we define the critical-layer set as the intersection of the two candidate sets:
\begin{equation}
\mathcal{L}^{\star}=\mathcal{L}_{\mathrm{sep}}\cap\mathcal{L}_{\mathrm{shift}}.
\end{equation}
The intersection retains layers consistently highlighted by both signals and provides a compact search space for subsequent LAN discovery. We perform all subsequent LAN discovery only within $\mathcal{L}^{\star}$, while the detailed statistical settings of $\tau_{\mathrm{sep}}$ and $\tau_{\mathrm{shift}}$ are given in the appendix.

\begin{table*}[t]
\centering
\tiny
\setlength{\tabcolsep}{3.0pt}
\renewcommand{\arraystretch}{1.04}
\arrayrulecolor{black}
\caption{\textbf{ACC and AP (\%) on MovieGen and Magic Videos.} Ours denotes V-FIND with the selected LAN subspace. Best results are in bold and second best are underlined.}
\label{tab:main-results}
\vspace{-0.3cm}
\resizebox{0.98\textwidth}{!}{
\begin{tabular}{l!{\color{black}\vrule width 0.45pt}cccccccccccc!{\color{black}\vrule width 0.45pt}cc!{\color{black}\vrule width 0.45pt}cc}
\hline\hline
\rowcolor{TableHeaderGray}
\textbf{Model} &
\multicolumn{2}{c}{\textbf{Wan 2.1}} &
\multicolumn{2}{c}{\textbf{Wan 1.3B}} &
\multicolumn{2}{c}{\textbf{Hailuo}} &
\multicolumn{2}{c}{\textbf{Seaweed}} &
\multicolumn{2}{c}{\textbf{Seedance}} &
\multicolumn{2}{c!{\color{black}\vrule width 0.45pt}}{\textbf{StepVideo}} &
\multicolumn{2}{c!{\color{black}\vrule width 0.45pt}}{\textbf{Mean}} &
\multicolumn{2}{c}{\textbf{MovieGen}} \\
\rowcolor{TableHeaderGray}
& \textbf{ACC} & \textbf{AP}
& \textbf{ACC} & \textbf{AP}
& \textbf{ACC} & \textbf{AP}
& \textbf{ACC} & \textbf{AP}
& \textbf{ACC} & \textbf{AP}
& \textbf{ACC} & \textbf{AP}
& \textbf{ACC} & \textbf{AP}
& \textbf{mACC} & \textbf{mAP} \\
\hline
RINE~\citep{koutlis2024leveraging} & 45.35 & 40.41 & 33.56 & 32.69 & 51.40 & 47.46 & 51.63 & 46.54 & 44.65 & 38.98 & 70.23 & 69.93 & 49.47 & 46.00 & 52.97 & 71.11 \\
FatFormer~\citep{liu2024fatformer} & 50.00 & 39.06 & 50.17 & 54.96 & 50.00 & 43.80 & 50.00 & 44.55 & 50.00 & 42.16 & 50.23 & 52.36 & 50.07 & 46.15 & 50.02 & 58.84 \\
B-Free~\citep{guillaro2025biasfree} & 53.72 & 59.80 & 68.32 & 73.81 & 55.81 & 62.62 & 31.63 & 36.08 & 36.05 & 38.93 & 48.60 & 52.37 & 49.02 & 53.94 & 64.30 & 70.38 \\
Effort~\citep{yan2025effort} & 73.26 & 76.61 & 66.95 & 73.75 & 67.44 & 70.67 & \underline{86.05} & 95.48 & 83.26 & 91.65 & 62.79 & 63.41 & 73.29 & 78.60 & 70.74 & 80.60 \\
WaveRep~\citep{corvi2025seeing} & 58.84 & 83.50 & 53.60 & 85.31 & 59.53 & 95.64 & 59.53 & \underline{99.12} & 57.21 & 78.64 & 59.53 & 94.04 & 58.04 & 89.38 & 65.30 & 92.97 \\
VINA~\citep{li2026vina} & 75.35 & 95.44 & \underline{89.90} & 98.08 & 74.19 & 94.47 & 73.02 & 90.51 & 74.42 & 95.00 & 75.58 & \textbf{97.09} & 77.08 & \underline{95.10} & 98.50 & 99.80 \\
F3Net~\citep{qian2020thinking} & 71.86 & 85.79 & 69.86 & 78.05 & 66.98 & 80.31 & 74.88 & 92.13 & 73.26 & 87.32 & 66.98 & 81.26 & 70.64 & 84.14 & 92.51 & 96.20 \\
TALL~\citep{xu2023tall} & 64.65 & 83.59 & 67.81 & 80.78 & 61.86 & 78.58 & 64.42 & 88.81 & 63.49 & 82.81 & 63.49 & 83.39 & 64.29 & 82.99 & 91.71 & 96.07 \\
NPR~\citep{tan2024rethinking} & 71.63 & 86.93 & 66.27 & 82.70 & 72.33 & 87.51 & 74.88 & 92.71 & 73.26 & 92.08 & 72.09 & 90.96 & 71.74 & 88.82 & 92.66 & 97.10 \\
TimeSformer~\citep{li2026preserving} & 68.84 & 82.44 & 72.09 & 83.04 & 66.28 & 75.42 & 69.77 & 83.49 & 67.91 & 80.29 & 67.21 & 78.55 & 68.68 & 80.54 & 91.41 & 96.91 \\
CLIP ViT-L/14~\citep{li2026preserving} & 78.14 & 98.78 & 77.74 & 93.14 & 77.21 & 87.43 & 77.91 & 97.58 & 77.21 & 92.05 & 75.58 & 86.85 & 77.30 & 92.64 & \textbf{99.20} & \textbf{99.95} \\
X-CLIP-B/16~\citep{li2026preserving} & 76.28 & 94.94 & 72.43 & 94.07 & 75.12 & \textbf{95.76} & 75.12 & 87.65 & 75.35 & \textbf{97.78} & 72.33 & 81.65 & 74.44 & 91.98 & 98.55 & 99.87 \\
X-CLIP-L/14~\citep{li2026preserving} & 80.00 & \textbf{99.62} & 86.13 & 82.73 & 79.53 & 92.91 & 78.60 & \textbf{99.16} & 80.00 & 96.18 & \underline{79.53} & \underline{95.71} & 80.63 & 94.39 & \underline{98.85} & \underline{99.94} \\
Moon-ViT~\citep{li2026preserving} & 76.74 & 93.84 & 79.62 & 90.39 & 75.81 & 87.24 & 76.74 & 94.11 & 75.81 & 90.68 & 74.88 & 81.69 & 76.60 & 89.66 & 98.25 & 99.24 \\
Qwen2.5-ViT~\citep{li2026preserving} & \underline{85.81} & 96.67 & 83.39 & \textbf{99.68} & \underline{84.65} & 94.20 & 83.95 & 91.59 & \underline{84.88} & 96.92 & 76.51 & 80.63 & \underline{83.20} & 93.28 & 97.20 & 99.46 \\
\hline
\rowcolor{TableAccentBlue}
\textbf{LANs (Ours)} &
\textbf{93.67} & \underline{99.34} &
\textbf{91.78} & \underline{99.40} &
\textbf{87.86} & \underline{95.70} &
\textbf{91.12} & 97.53 &
\textbf{89.95} & \underline{97.23} &
\textbf{84.19} & 93.02 &
\textbf{89.76}$\pm$0.28 & \textbf{97.04}$\pm$0.14 &
96.86$\pm$0.20 & 99.30$\pm$0.05 \\
\hline\hline
\end{tabular}
}
\vspace{-0.3cm}
\end{table*}

\subsection{Latent Anchor Neuron Discovery}
After obtaining the critical-layer set $\mathcal{L}^{\star}$, we further identify sparse latent anchor neurons within these layers. Let the full candidate neuron set in the critical layers be
\begin{equation}
\mathcal{N}^{\star}=\{(\ell,k)\mid \ell\in\mathcal{L}^{\star},\,1\leq k\leq M\},
\end{equation}
where $k$ indexes the latent unit in the unit-level representation of layer $\ell$. Since latent units may differ in scale across layers, we thus score them separately within each critical layer. For each $\ell\in\mathcal{L}^{\star}$, we in turn train a linear probe on the pooled representation:
\begin{equation}
g^{(\ell)}(x)={\mathbf{w}^{(\ell)}}^\top\mathbf{a}^{(\ell)}(x)+b^{(\ell)}.
\end{equation}
Each probe is independently trained on the discovery split using a standard cross-entropy loss for real--fake classification, with its weights characterizing the discriminative alignment of individual units.

For a candidate neuron $n=(\ell,k)$, let its corresponding probe coefficient be $w_n=w_k^{(\ell)}$. For an input video $x_i$, we define the per-sample response of this neuron as
\begin{equation}
r_n(x_i)=|a_n(x_i)|\cdot |w_n|.
\end{equation}
This response combines two factors: the activation magnitude of the neuron on the current sample and the discriminative weight assigned to that neuron by the layer-specific probe. It therefore provides a unified unit-level response measure for subsequent selection.

Per-sample response alone is insufficient to determine whether a neuron can separate real and fake videos consistently, so we further measure the response separation between the two groups. For class $c\in\{\mathrm{real},\mathrm{fake}\}$, we define the mean response of neuron $n$ as
\begin{equation}
\mu_{n,c}=\frac{1}{|\mathcal{D}_c|}\sum_{x\in\mathcal{D}_c} r_n(x),
\end{equation}
where $\sigma_n^{\mathrm{pool}}$ denotes the pooled within-group standard deviation of the per-neuron response, with its exact definition given in the appendix. It serves to normalize the response gap by the within-group variation, and $\epsilon$ is used to prevent division by zero. Based on these statistics, we define the separation effect size of neuron $n$ as
\begin{equation}
d_n=\frac{|\mu_{n,\mathrm{fake}}-\mu_{n,\mathrm{real}}|}{\sigma_n^{\mathrm{pool}}+ \epsilon}.
\end{equation}
We then use an effect-size threshold $\tau_d$ as the selection criterion and retain
\begin{equation}
\mathcal{S}=\{n\in\mathcal{N}^{\star}\mid d_n\geq\tau_d\},
\end{equation}
as the final LANs. Here $\tau_d$ is the core hyperparameter in the neuron-selection stage, and its effect is analyzed in the ablation study.

After obtaining the selected LAN set $\mathcal{S}$, we use their raw activations in the frozen backbone to construct the final feature space, rather than the discovery-stage response scores $r_n(x)$. Specifically, we concatenate these activations into a compact forensic subspace
\begin{equation}
\mathbf{h}_{\mathcal{S}}(x)=\left[a_n(x)\right]_{n\in\mathcal{S}},
\end{equation}
and train a lightweight linear classifier on top of it while keeping the original detector backbone frozen.

\section{Experiments and Analysis}

\noindent\textbf{Datasets.} We strictly separate neuron discovery from external evaluation throughout all experiments. The original detector is trained on the 15Model-140K dataset~\citep{li2026preserving}. From a smaller subset of this source distribution, we construct mutually disjoint discovery, training, and validation splits for layer localization, LAN discovery, and final linear-readout training. External evaluation is conducted on three benchmarks: Magic Videos~\citep{li2026preserving}, MovieGen~\citep{polyak2024moviegen}, and DVF~\citep{song2024multimodal}. Among them, Magic Videos focuses on recent high-quality generated videos, while MovieGen and DVF test generalization beyond the source dataset.

\noindent\textbf{Metrics.} Following the evaluation conventions of DeepfakeBench~\citep{yan2023deepfakebench} and the Magic protocol used by \citet{li2026preserving}, we report mean ACC (mACC) and mean AP (mAP) for Magic Videos and MovieGen, and mean AUC for DVF in our evaluation. ACC measures standard binary classification accuracy and has been adopted in deepfake and generated-video detection benchmarks~\citep{yan2023deepfakebench,ni2025genvidbench}; AP evaluates global precision--recall ranking quality and is reported in these benchmarks~\citep{yan2023deepfakebench,ni2025genvidbench,li2026preserving}; AUC summarizes detector performance across all decision thresholds and is widely used for DVF-style evaluation~\citep{yan2023deepfakebench,song2024multimodal,li2026preserving}.

\noindent\textbf{Hyperparameters.} Unless otherwise specified, we use the released Qwen2.5-ViT detector~\citep{li2026preserving} as the frozen backbone and train only a lightweight linear classifier on the selected LAN subspace. In the main setting, we use the critical layers $\{28,29,30\}$ and select neurons solely by the discriminative effect size $d_n$ with the default threshold $\tau_d=1.5$, yielding 211 LANs. The discovery, training, and validation splits contain 2400, 2400, and 1200 videos, respectively. Each video is decoded at 2 FPS, and the center-aligned 8 frames are fed into the Qwen2.5-ViT visual encoder with dynamic resolution from 224p to 720p while preserving aspect ratio. We optimize both the layer-wise probes and the final linear classifier with AdamW using a learning rate of $1\times10^{-2}$ and weight decay of $1\times10^{-4}$; the former is trained for 300 epochs and the latter for 200 epochs.

\subsection{Compare with SOTA Methods}
 We compare against four categories of baselines: AI-generated video detectors, including MM-Det~\citep{song2024multimodal}, VINA~\citep{li2026vina}, and WaveRep~\citep{corvi2025seeing}; visual and video foundation backbones, including CLIP~\citep{radford2021learning}, X-CLIP~\citep{ni2022expanding}, TimeSformer~\citep{bertasius2021space}, Moon-ViT~\citep{du2025kimi}, and Qwen2.5-ViT~\citep{li2026preserving}; deepfake detectors, including F3Net~\citep{qian2020thinking} and TALL~\citep{xu2023tall}; and general AI-generated image detectors, including CNNDet~\citep{wang2020cnn}, DIRE~\citep{wang2023dire}, NPR~\citep{tan2024rethinking}, RINE~\citep{koutlis2024leveraging}, FatFormer~\citep{liu2024fatformer}, B-Free~\citep{guillaro2025biasfree}, and Effort~\citep{yan2025effort}.

\noindent\textbf{MovieGen and Magic Videos.}
To verify whether the discovered forensic subspace supports strong detection across external benchmarks for generated videos, we conduct comparative experiments against state-of-the-art detectors on MovieGen and Magic Videos. Averaged over five seeds, V-FIND achieves 89.76 mACC and 97.04 mAP on Magic Videos (Table~\ref{tab:main-results}), improving the original Qwen2.5-ViT detector by 6.56 and 3.76 points, respectively. ACC improves on all six Magic subsets, with gains ranging from 3.21 points on Hailuo to 8.39 points on Wan 1.3B. AP improves on five subsets, with the largest gain on StepVideo (+12.39 points), and decreases only slightly on Wan 1.3B (-0.28 points). On MovieGen, V-FIND obtains 96.86 ACC and 99.30 AP, which is slightly below the original detector (97.20/99.46) but remains competitive with the other compared methods. The discovered forensic subspace contains reusable forgery-discriminative knowledge that supports strong detection without full-model retraining.

\noindent\textbf{DVF Benchmark.}
For DVF, V-FIND obtains an average AUC of 98.2 (Table~\ref{tab:dvf-auc}), 0.7 points above the original Qwen2.5-ViT detector (97.5) and the best average among the compared methods. At the subset level, it improves over the original detector on OpenSora (+1.4), Sora (+1.5), Pika (+0.8), Stable Diffusion (+1.5), and Stable Video (+0.2), whereas it decreases slightly on VideoCrafter1 (99.4 vs. 99.8) and ZeroScope (92.8 vs. 93.5). This result indicates that the extracted forgery-discriminative information remains effective on an external benchmark, supporting cross-dataset applicability of the discovered subspace.

\begin{table}[t]
\centering
\scriptsize
\setlength{\tabcolsep}{1.9pt}
\renewcommand{\arraystretch}{1.08}
\arrayrulecolor{black}
\caption{\textbf{Benchmarking Results (AUC, \%) on DVF-Test.} Ours denotes V-FIND using the selected LAN subspace. Best results are in \textbf{bold} and second-best are \underline{underlined}.}
\label{tab:dvf-auc}
\vspace{-0.3cm}
\resizebox{0.98\columnwidth}{!}{
\begin{tabular}{l!{\color{black}\vrule width 0.45pt}cccccccc}
\hline\hline
\cellcolor{TableHeaderGray}\textbf{Method} &
\cellcolor{TableHeaderGray}\textbf{\shortstack{Video\\Crafter1}} &
\cellcolor{TableHeaderGray}\textbf{\shortstack{Zero\\Scope}} &
\cellcolor{TableHeaderGray}\textbf{\shortstack{Open\\Sora}} &
\cellcolor{TableHeaderGray}\textbf{Sora} &
\cellcolor{TableHeaderGray}\textbf{Pika} &
\cellcolor{TableHeaderGray}\textbf{\shortstack{Stable\\Diff.}} &
\cellcolor{TableHeaderGray}\textbf{\shortstack{Stable\\Video}} &
\cellcolor{TableHeaderGray}\textbf{AVG} \\
\hline
CNNDet~\citep{wang2020cnn} & 87.4 & 88.2 & 78.0 & 63.8 & 77.3 & 73.5 & 78.9 & 78.2 \\
DIRE~\citep{wang2023dire} & 55.9 & 61.8 & 53.8 & 60.5 & 65.8 & 62.7 & 69.9 & 61.5 \\
MM-Det~\citep{song2024multimodal} & 93.5 & \textbf{94.0} & 88.8 & 86.2 & 95.9 & \underline{95.7} & 89.9 & 92.0 \\
NPR~\citep{tan2024rethinking} & 86.6 & 85.6 & 96.0 & 81.0 & 94.6 & 71.1 & 97.0 & 87.4 \\
TALL~\citep{xu2023tall} & 95.4 & 91.8 & 97.2 & 94.9 & 97.5 & 83.6 & 98.2 & 94.1 \\
F3Net~\citep{qian2020thinking} & 90.4 & 90.2 & 95.9 & 90.1 & 97.8 & 93.1 & 98.5 & 93.7 \\
TimeSformer~\citep{bertasius2021space} & 94.5 & 92.7 & 98.0 & 92.5 & 98.4 & 92.4 & 99.5 & 95.4 \\
Qwen2.5-ViT~\citep{li2026preserving} & \textbf{99.8} & \underline{93.5} & \underline{98.6} & \underline{96.4} & \underline{99.1} & 95.6 & \underline{99.7} & \underline{97.5} \\
\hline
\cellcolor{TableAccentBlue}\textbf{LANs (Ours)} &
\cellcolor{TableAccentBlue}\underline{99.4} &
\cellcolor{TableAccentBlue}92.8 &
\cellcolor{TableAccentBlue}\textbf{100.0} &
\cellcolor{TableAccentBlue}\textbf{97.9} &
\cellcolor{TableAccentBlue}\textbf{99.9} &
\cellcolor{TableAccentBlue}\textbf{97.1} &
\cellcolor{TableAccentBlue}\textbf{99.9} &
\cellcolor{TableAccentBlue}\textbf{98.2}$\pm$0.1 \\
\hline\hline
\end{tabular}
}
\vspace{-0.4cm}
\end{table}

\begin{figure}[t]
\centering
\includegraphics[width=0.99\linewidth]{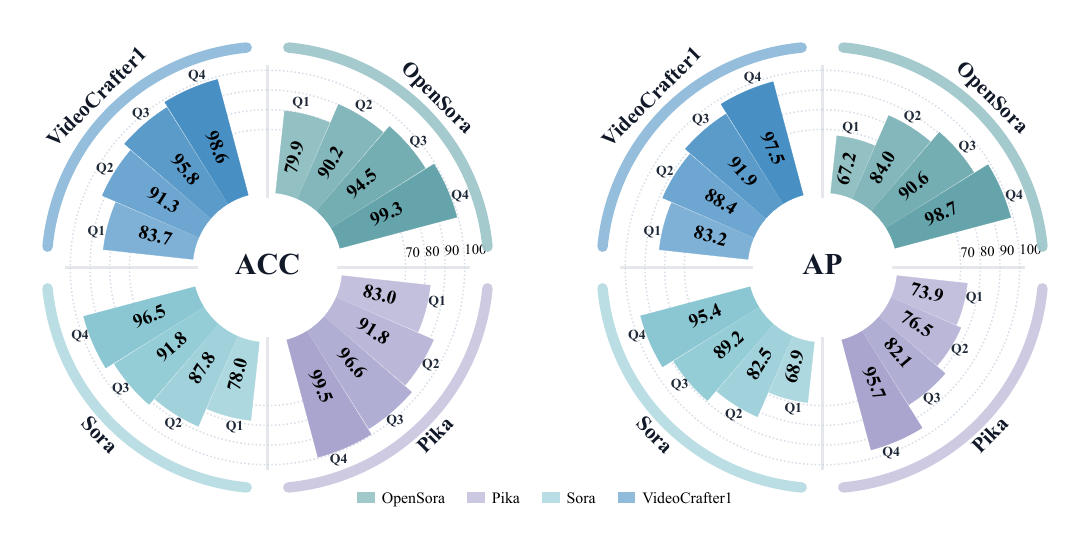}
\vspace{-0.3cm}
\caption{\textbf{Layer Localization.} Quarter-level ACC and AP on four DVF subsets, where Q1--Q4 denote layers 0--7, 8--15, 16--23, and 24--31, respectively.}
\label{fig:depth-wise-validation}
\vspace{-0.4cm}
\end{figure}

\subsection{Ablation Study}

\noindent\textbf{Layer Localization.}
To verify whether the critical layers carrying forgery-discriminative signals are located in the late part of the backbone, we conduct a depth-interval comparison by dividing the 32 layers into Q1 (0--7), Q2 (8--15), Q3 (16--23), and Q4 (24--31) under the same probing protocol. The results show that ACC and AP generally increase with layer depth across four representative DVF subsets, with Q4 consistently achieving the strongest performance, as shown in Figure~\ref{fig:depth-wise-validation}. Relative to Q1, Q4 improves ACC by 14.9--19.5 and AP by 14.2--31.6 points; for example, OpenSora increases from 79.89/67.15 to 99.34/98.74 and VideoCrafter1 from 83.72/83.25 to 98.65/97.46. This trend is consistent with previous observations on temporal coherence and inconsistency cues \citep{guo2025tfcu} and with the role of deeper visual or video transformer layers in encoding more integrated temporal information \citep{chen2025shallower,shi2026causality}. This consistency supports the layer-localization choice: the late-layer region provides a more favorable representation for the subsequent forgery readout. Therefore, these results support focusing LAN discovery on this region and treating the intersection of the two discrepancy signals as a practical candidate-selection criterion.

\begin{figure}[t]
\centering
\includegraphics[width=0.99\linewidth]{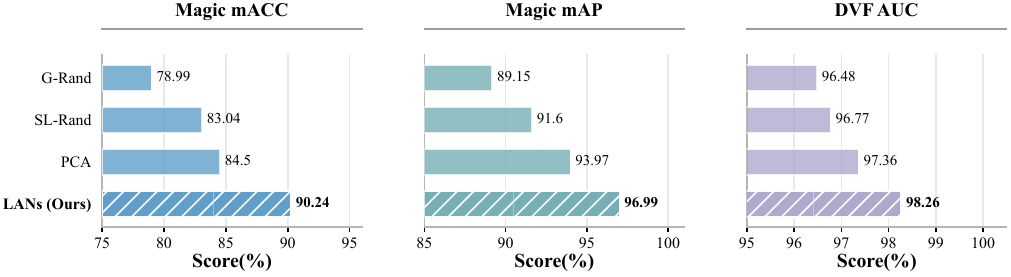}
\vspace{-0.3cm}
\caption{\textbf{Neuron Selection.} We compare 211-dimensional fixed-budget controls. G-Rand samples across the backbone, whereas SL-Rand preserves the LAN layer allocation.}
\label{fig:fixed-budget-controls}
\end{figure}

\noindent\textbf{Neuron Selection.}
To verify that LANs carry genuinely forgery-discriminative signals rather than simply benefiting from a constrained feature dimension, we evaluate them against several fixed-budget baselines in Figure~\ref{fig:fixed-budget-controls}. The comparison reveals two clear takeaways: first, layer localization isolates a more informative search space; second, LAN discovery successfully extracts a highly effective forensic subspace within these layers. Specifically, restricting random selection to localized layers improves Magic mACC/mAP from 78.99/89.15 to 83.04/91.60 and DVF AUC from 96.48 to 96.77. Under the same 211-dimensional budget, LANs achieve superior performance (90.24\% mACC, 96.99\% mAP, and 98.26\% AUC), consistently outperforming both Same-layer Random and PCA. These results indicate that the performance gains stem from isolating functionally specialized neurons that encode more forgery-discriminative knowledge, rather than feature compactness alone.

\begin{figure}[t]
\centering
\includegraphics[width=0.99\linewidth]{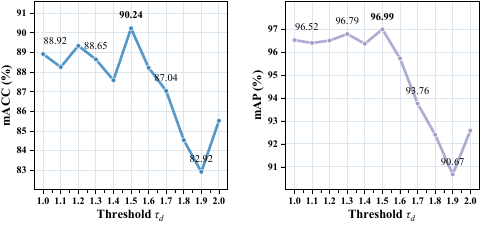}
\vspace{-0.30cm}
\caption{\textbf{Threshold Sensitivity.} A coarse-to-fine sweep is used to select the default setting $\tau_d=1.5$.}
\label{fig:threshold-sensitivity}
\end{figure}

\begin{figure*}[t]
\centering
\includegraphics[width=0.98\linewidth]
{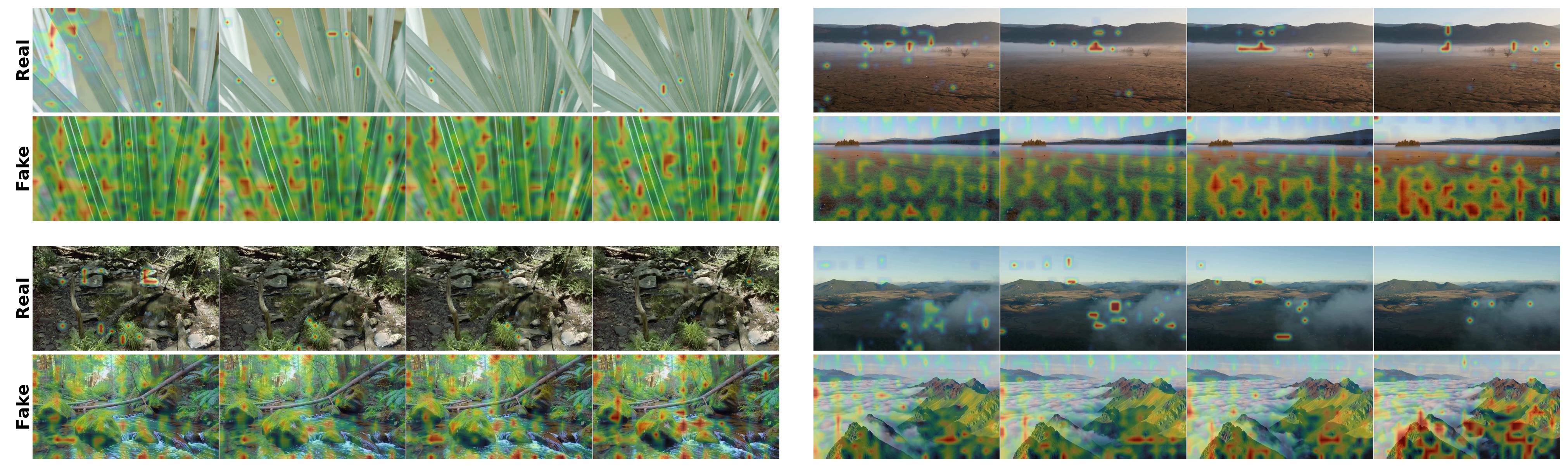}
\vspace{-0.30cm}
\caption{\textbf{Visualization of LANs Heatmaps.}
Heatmaps of latent anchor neurons (LANs) are shown for paired real and forged videos from the Magic Videos, with consecutive frames arranged from left to right.}
\label{fig:lan-qualitative}
\vspace{-0.4cm}
\end{figure*}

\noindent\textbf{Threshold Sensitivity.}
To verify whether the threshold for identifying LANs can be selected in a systematic and stable manner rather than being heuristically fixed, we conduct a coarse-to-fine sweep over $\tau_d$. The results identify a relatively stable and consistently high-performing region around $\tau_d \approx 1.5$, as shown in Figure~\ref{fig:threshold-sensitivity}. Specifically, we evaluate detection performance under different threshold settings, and the results show that $\tau_d=1.5$ achieves the best trade-off between accuracy and average precision, reaching 90.24\% mACC and 96.99\% mAP. Nearby settings remain competitive, with 88.22/95.72 at $\tau_d=1.6$ and 87.59/96.36 at $\tau_d=1.4$. Within the plotted 1.0--2.0 range, the lowest values occur at $\tau_d=1.9$ (82.92/90.67), while $\tau_d=2.0$ and $\tau_d=1.2$ reach 85.52/92.58 and 89.35/96.50, respectively. Therefore, we select $\tau_d=1.5$ based on the observed sensitivity behavior rather than an arbitrary heuristic choice.

\subsection{Deeper Analysis}

\begin{table}[t]
\centering
\small
\setlength{\tabcolsep}{2.6pt}
\renewcommand{\arraystretch}{1.15}
\arrayrulecolor{black}
\caption{\textbf{Functional Specificity.} This table reports mean decision-logit shifts ($\Delta$) and label-flip rates under LAN and budget-matched random interventions. Flip counts are reported over eligible correctly classified pairs.}
\label{tab:causal-validation}
\vspace{-0.3cm}
\begin{tabular}{c!{\color{black}\vrule width 0.45pt}cccc}
\hline\hline
\cellcolor{TableHeaderGray} &
\multicolumn{2}{c}{\cellcolor{TableHeaderGray}\textbf{Random}} &
\multicolumn{2}{c}{\cellcolor{TableHeaderGray}\textbf{LANs (Ours)}} \\
\cline{2-5}
\multirow{-2}{*}{\cellcolor{TableHeaderGray}\textbf{Setting}} &
\cellcolor{TableHeaderGray}\textbf{$\Delta$ Logit} &
\cellcolor{TableHeaderGray}\textbf{Flip Rate (\%)} &
\cellcolor{TableHeaderGray}\textbf{$\Delta$ Logit} &
\cellcolor{TableHeaderGray}\textbf{Flip Rate (\%)} \\
\hline
Fake$\rightarrow$Real & +0.414 & 6.8 \,(66/970) & \cellcolor{TableAccentBlue}\textbf{+0.884} & \cellcolor{TableAccentBlue}\textbf{19.7} \,(\textbf{191}/\textbf{970}) \\
Real$\rightarrow$Fake & $-0.282$ & 1.2 \,(15/1304) & \cellcolor{TableAccentBlue}$\mathbf{-0.730}$ & \cellcolor{TableAccentBlue}\textbf{1.8} \,(\textbf{23}/\textbf{1304}) \\
\hline\hline
\end{tabular}
\vspace{-0.4cm}
\end{table}
\noindent\textbf{Neuron Responses.}
To verify whether the discovered LANs consistently carry forgery-discriminative signals, we visualize their heatmaps on paired real and forged videos from Magic. Figure~\ref{fig:lan-qualitative} shows both the response differences between real and fake videos and the activation patterns across consecutive frames. Compared with real videos, fake videos produce stronger, broader, and more continuous LAN responses around complex textures, object boundaries, repeated structures, and local visual anomalies. These responses also remain relatively stable across adjacent frames. In contrast, real videos show weaker and sparser activations. These observations suggest that LANs focus on local forgery cues. Within the forgery detector, latent anchor neurons can be organized into a compact forensic subspace, enabling the extraction of the forgery-discriminative knowledge.

\noindent\textbf{Functional Specificity.}
To verify that the discovered LANs carry dedicated forgery-discriminative signals rather than arbitrary activations, we perform directional activation-swap interventions on paired real and synthetic videos, comparing them against budget-matched, same-layer random controls (Table~\ref{tab:causal-validation}). Intervening on LANs induces substantially larger target-logit shifts and higher directional label-flip rates across both transfer directions. Specifically, LAN activation swaps drive logit shifts of $+0.884$ (vs. $+0.414$ for random controls) in Fake$\rightarrow$Real transfers and $-0.730$ (vs. $-0.282$) in Real$\rightarrow$Fake transfers, and increase flip rates from 6.8\% to 19.7\% and 1.2\% to 1.8\%, respectively. These causal intervention results provide direct evidence that LANs exhibit clear functional specificity for video forgery detection.

\begin{table}[t]
\centering
\small
\renewcommand{\arraystretch}{1.10}
\setlength{\tabcolsep}{5.5pt}
\arrayrulecolor{black}
\caption{\textbf{Cross-Architecture Validation.} We compare each original detector with its LAN-subspace classifier on Magic; parentheses denote changes over the original detector.}
\label{tab:cross-architecture-magic}
\vspace{-0.3cm}
\begin{tabular}{l!{\color{black}\vrule width 0.45pt}cc}
\hline\hline
\multicolumn{1}{l!{\color{black}\vrule width 0.45pt}}{\cellcolor{TableHeaderGray}\textbf{Model}} &
\multicolumn{1}{c}{\cellcolor{TableHeaderGray}\textbf{mACC} ($\uparrow$)} &
\multicolumn{1}{c}{\cellcolor{TableHeaderGray}\textbf{mAP} ($\uparrow$)} \\
\hline
\textcolor{gray}{VINA~\citep{li2026vina}} & \textcolor{gray}{77.08} & \textcolor{gray}{95.10} \\
\multicolumn{1}{l!{\color{black}\vrule width 0.45pt}}{\cellcolor{TableAccentBlue}\textbf{+ LANs (Ours)}} &
\multicolumn{1}{c}{\cellcolor{TableAccentBlue}\textbf{84.65}\ {\scriptsize\textcolor{red}{(+7.6)}}} &
\multicolumn{1}{c}{\cellcolor{TableAccentBlue}94.84\ {\scriptsize\textcolor{gray}{(-0.3)}}} \\
\hline
\textcolor{gray}{X-CLIP-B/16~\citep{li2026preserving}} & \textcolor{gray}{74.44} & \textcolor{gray}{91.98} \\
\multicolumn{1}{l!{\color{black}\vrule width 0.45pt}}{\cellcolor{TableAccentBlue}\textbf{+ LANs (Ours)}} &
\multicolumn{1}{c}{\cellcolor{TableAccentBlue}\textbf{89.29}\ {\scriptsize\textcolor{red}{(+14.9)}}} &
\multicolumn{1}{c}{\cellcolor{TableAccentBlue}\textbf{96.16}\ {\scriptsize\textcolor{red}{(+4.2)}}} \\
\hline
\textcolor{gray}{X-CLIP-L/14~\citep{li2026preserving}} & \textcolor{gray}{80.63} & \textcolor{gray}{94.39} \\
\multicolumn{1}{l!{\color{black}\vrule width 0.45pt}}{\cellcolor{TableAccentBlue}\textbf{+ LANs (Ours)}} &
\multicolumn{1}{c}{\cellcolor{TableAccentBlue}\textbf{89.08}\ {\scriptsize\textcolor{red}{(+8.5)}}} &
\multicolumn{1}{c}{\cellcolor{TableAccentBlue}\textbf{96.97}\ {\scriptsize\textcolor{red}{(+2.6)}}} \\
\hline\hline
\end{tabular}
\end{table}

\noindent\textbf{Cross-Architecture Validation.} To examine whether the same LAN-discovery pipeline can be applied beyond Qwen2.5-ViT, we apply it to three additional backbones: VINA, X-CLIP-B/16, and X-CLIP-L/14, and evaluate the resulting forensic subspaces on Magic Videos (Table~\ref{tab:cross-architecture-magic}). The discovered subspaces improve mACC across all three detectors, while boosting mAP on both X-CLIP variants and preserving competitive mAP on VINA. Specifically, mACC improves by 7.57, 14.85, and 8.45 percentage points on VINA, X-CLIP-B/16, and X-CLIP-L/14, respectively. Concretely, mACC rises from 77.08 to 84.65 on VINA, from 74.44 to 89.29 on X-CLIP-B/16, and from 80.63 to 89.08 on X-CLIP-L/14; their mAP changes from 95.10 to 94.84, from 91.98 to 96.16, and from 94.39 to 96.97, respectively. These results indicate that the LAN-subspace readout remains effective across the three tested backbones.

\begin{figure}[t]
\centering
\includegraphics[width=0.944\linewidth]{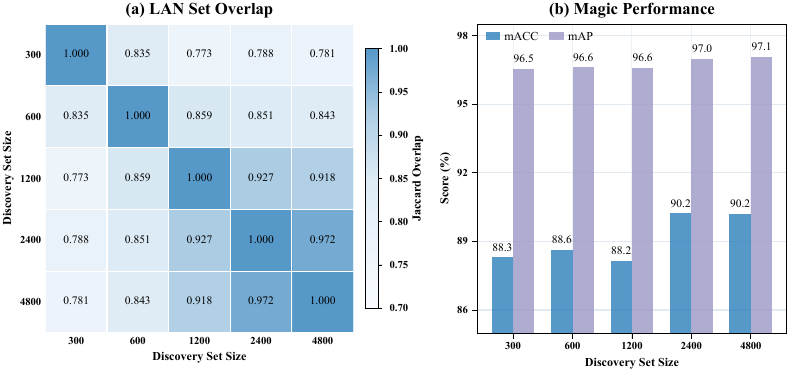}
\vspace{-0.3cm}
\caption{\textbf{Discovery Stability.} \textbf{(a)} Pairwise Jaccard overlap between LAN sets discovered from pools ranging from 300 to 4800 videos. \textbf{(b)} Magic mACC and mAP (\%) for the corresponding 211-channel LAN subspaces. All settings use the same channel budget and classifier protocol.}
\label{fig:discover-data-size}
\vspace{-0.3cm}
\end{figure}

\noindent\textbf{Discovery Stability.}
To verify the robustness of LAN discovery to the amount of discovery data, we vary the discovery pool over 300, 600, 1200, 2400, and 4800 videos while keeping the selected-channel budget, train/val split, and final classifier identical. Figure~\ref{fig:discover-data-size}(a) reports the pairwise Jaccard overlap, and Figure~\ref{fig:discover-data-size}(b) reports the corresponding Magic performance. Figure~\ref{fig:discover-data-size}(a) shows that overlap is generally higher between neighboring pool sizes, indicating that larger discovery sets select increasingly aligned LAN sets; across adjacent discovery-set sizes, the Jaccard overlap increases from 0.835 to 0.972. Figure~\ref{fig:discover-data-size}(b) shows that the corresponding Magic performance remains stable across all pool sizes, with mACC and mAP varying by at most 2.1 and 0.6 points, respectively. This behavior suggests that adding discovery videos mainly refines the same neuron population rather than replacing it. Overall, these results indicate that LAN discovery and the resulting forensic subspace exhibit a certain degree of robustness to the discovery-set size.

\noindent\textbf{Readout Data Efficiency.}
To verify whether the compact forensic subspace can support detection with limited training data, we evaluate V-FIND under different training-data sizes while keeping the original backbone frozen and training only a lightweight linear classifier. The results show that performance peaks around 2400 samples and remains relatively stable afterward, as shown in Table~\ref{tab:train-size-robustness}. Specifically, on Seaweed, accuracy increases from 84.65\% with 300 samples to 91.86\% with 2400 samples, before changing to 89.07\% with 4800 samples. Across all six Magic subsets, the mean ACC is 85.9\% at 300 samples, 84.3\% at 600, 87.4\% at 1200, 90.2\% at 2400, and 88.6\% at 4800. Overall, the mean ACC first decreases from 300 to 600 samples, then increases steadily through 1200 to its peak at 2400, before declining modestly at 4800. Therefore, these results suggest that the discovered forensic subspace can support strong detection with a relatively limited amount of training data in this evaluation setting.

\begin{table}[t]
\centering
\footnotesize
\setlength{\tabcolsep}{2.0pt}
\renewcommand{\arraystretch}{1.12}
\arrayrulecolor{black}
\caption{\textbf{Readout Data Efficiency.} ACC (\%) under different training-data sizes across Magic Video models using the compact LAN subspace.}
\label{tab:train-size-robustness}
\vspace{-0.3cm}
\resizebox{\columnwidth}{!}{
\begin{tabular}{l!{\color{black}\vrule width 0.45pt}cccccc!{\color{black}\vrule width 0.45pt}c}
\hline\hline
\cellcolor{TableHeaderGray}\textbf{Size} &
\cellcolor{TableHeaderGray}\textbf{StepVideo} &
\cellcolor{TableHeaderGray}\textbf{Seaweed} &
\cellcolor{TableHeaderGray}\textbf{Hailuo} &
\cellcolor{TableHeaderGray}\textbf{Seedance} &
\cellcolor{TableHeaderGray}\textbf{Wan 2.1} &
\cellcolor{TableHeaderGray}\textbf{Wan 1.3B} &
\cellcolor{TableHeaderGray}\textbf{Mean ACC} \\
\hline
300 & 76.7 & 84.7 & 84.2 & 86.3 & 90.5 & 93.2 & 85.9 \\
600 & 72.1 & 82.6 & 81.6 & 84.2 & 90.9 & 94.3 & 84.3 \\
1200 & 78.1 & 87.2 & 83.7 & 87.4 & 93.5 & 94.2 & 87.4 \\
\textbf{2400} & \textbf{84.9} & \textbf{91.9} & \textbf{88.1} & \textbf{90.2} & \textbf{94.2} & \textbf{92.1} & \textbf{90.2} \\
4800 & 83.5 & 89.1 & 87.2 & 88.6 & 91.9 & 91.4 & 88.6 \\
\hline\hline
\end{tabular}
}
\vspace{-0.3cm}
\end{table}

\begin{figure}[t]
\centering
\includegraphics[width=0.84\linewidth]{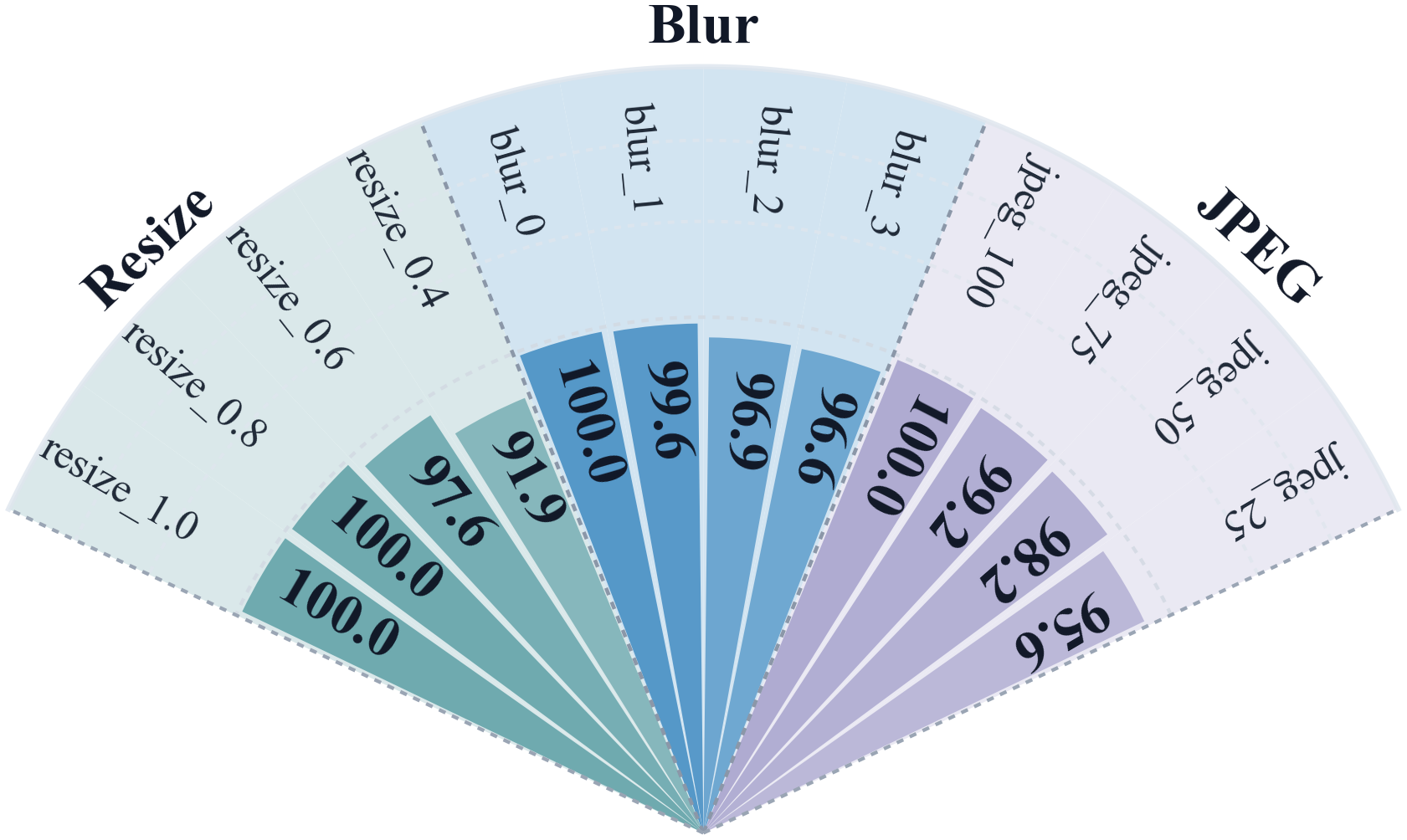}
\vspace{-0.3cm}
\caption{\textbf{Input Robustness.} Relative mAP (\%) under common input perturbations (resizing, blur, and JPEG compression) using the compact LAN subspace.}
\label{fig:robustness-perturbation}
\vspace{-0.4cm}
\end{figure}

\noindent\textbf{Input Robustness.}
To verify whether the discovered forensic subspace remains effective under common input degradations, we evaluate it on Magic Videos with different levels of Gaussian blur, JPEG compression, and resizing. The results show that the relative mAP remains close to the clean baseline under mild perturbations and gradually decreases as perturbation strength increases, as shown in Figure~\ref{fig:robustness-perturbation}. Specifically, the relative mAP reaches 99.62 under blur\_1, 99.23 under jpeg\_75, and 100.01 under resize\_0.8 at the mildest levels. Under stronger perturbations, the relative mAP further decreases to 96.91/96.57 at blur\_2/blur\_3, 98.20/95.62 at jpeg\_50/jpeg\_25, and 97.56/91.94 at resize\_0.6/resize\_0.4. Across all evaluated perturbation settings, the relative mAP remains above 91.9. Therefore, these results suggest that the forgery-discriminative signals retained in the compact forensic subspace remain effective under the evaluated mild degradations.

\section{Conclusion}
This paper studies video forgery detection from an intrinsic-knowledge perspective and proposes V-FIND, a video forgery-intrinsic neuron discovery framework. V-FIND localizes critical layers with pronounced real–fake discrepancies, identifies latent anchor neurons carrying forgery-discriminative signals, and organizes them into a compact forensic subspace. With the original backbone frozen and only a lightweight linear classifier trained, this subspace delivers strong detection performance across multiple external benchmarks, while neuron intervention experiments provide direct evidence for the functional specificity of the discovered neurons. Overall, our results suggest that a sparse set of functionally specialized neurons encodes more forgery-discriminative knowledge, indicating that a compact subset of neurons can provide an extractable and reusable forensic readout. Extending this neuron-level perspective to broader detector architectures and longer temporal settings remains a direction for future work.

\section*{LLM Usage Statement}
We used DeepSeek-V4.1-Flash through the Codex harness (codex-cli) to assist with language editing, clarity improvements, and data-processing tasks. The authors reviewed all resulting textual changes to confirm that they improved grammar, clarity, and readability without altering the intended presentation of the methods, results, or conclusions. The nature and extent of these uses were assistive only. The authors remain fully responsible for the research, all scientific claims, and the final manuscript.

{\small
\bibliographystyle{ieeenat_fullname}
\bibliography{main}
}

\clearpage
\noindent The appendix is organized as follows. Appendix~\ref{ap:notation} summarizes the notation used throughout the main paper and this appendix. Appendix~\ref{ap1} clarifies the experimental protocol and internal data-split usage. Appendix~\ref{ap2} provides deferred mathematical definitions for layer localization and LAN discovery. Appendix~\ref{ap3} presents a theoretical interpretation of layer localization and sparse LAN readout. Appendix~\ref{ap4} reports additional sensitivity analyses of the selected LAN subspace. Finally, Appendices~\ref{ap5} and~\ref{ap6} discuss methodological design choices, limitations, and future research directions.

\appendix
\section{Notation Summary}
\label{ap:notation}
Table~\ref{tab:notation} summarizes the core notation used to describe layer localization, latent anchor neuron (LAN) discovery, and the compact forensic subspace. A superscript $(\ell)$ indexes a model layer and does not denote exponentiation.

\begin{table*}[t]
\centering
\footnotesize
\setlength{\tabcolsep}{4pt}
\renewcommand{\arraystretch}{1.08}
\caption{\textbf{Notation Summary.} Core notation used in the main paper and appendix.}
\label{tab:notation}
\begin{tabular}{@{}p{0.23\linewidth}p{0.69\linewidth}@{}}
\hline
\textbf{Symbol} & \textbf{Definition} \\
\hline
$x=\{I_t\}_{t=1}^{T}$; $T,H,W,C$ & Input video clip, its sampled frames, frame height, frame width, and number of channels, respectively. \\
$H^{(\ell)}(x)$; $N_t,N_s,M$ & Hidden representation at layer $\ell$, with temporal-token, spatial-token, and latent-unit dimensions. \\
$\mathbf{a}^{(\ell)}(x)$; $a_k^{(\ell)}(x)$; $a_n(x)$ & Spatiotemporally pooled layer representation, the response of latent unit $k$, and the scalar activation associated with neuron $n$. \\
$n=(\ell,k)$ & Candidate latent anchor neuron indexed by layer $\ell$ and latent unit $k$. \\
$\mathcal{D}_c$; $c\in\{\mathrm{real},\mathrm{fake}\}$ & Class-conditional video set and the corresponding real or fake class label. \\
$\mu_c^{(\ell)}$ & Mean pooled representation of class $c$ at layer $\ell$. \\
$D_{\cos}(\ell)$; $\tau_{\mathrm{sep}}$; $\mathcal{L}_{\mathrm{sep}}$ & Directional centroid-separation score, its threshold, and the resulting candidate layer set. \\
$D_{\mathrm{shift}}(\ell)$; $\bar{\sigma}_{\ell}^{2}$; $\tau_{\mathrm{shift}}$; $\mathcal{L}_{\mathrm{shift}}$ & Normalized centroid-shift score, mean within-class channel variance, its threshold, and the resulting candidate layer set. \\
\hline
$\mathcal{L}^{\star}$ & Critical-layer set obtained by intersecting $\mathcal{L}_{\mathrm{sep}}$ and $\mathcal{L}_{\mathrm{shift}}$. \\
$\mathcal{N}^{\star}$ & Candidate neuron set formed by all latent units in the critical layers. \\
$g^{(\ell)}(x)$; $\mathbf{w}^{(\ell)}$; $b^{(\ell)}$ & Layer-specific linear probe and its weight vector and bias. \\
$w_n=w_k^{(\ell)}$ & Probe coefficient associated with candidate neuron $n=(\ell,k)$. \\
$r_n(x)$ & Probe-weighted per-sample response of neuron $n$. \\
$\mu_{n,c}$ & Mean response of neuron $n$ for class $c$. \\
$\sigma_{n,c}^{2}$ & Weight-normalized within-class response variance of neuron $n$. \\
$\sigma_n^{\mathrm{pool}}$ & Pooled within-group standard deviation used to normalize the response gap. \\
$\epsilon$ & Small positive constant that prevents division by zero. \\
$d_n$ & Separation effect size of neuron $n$. \\
$\tau_d$; $\mathcal{S}$ & Effect-size threshold and the selected LAN set. \\
\hline
$\mathbf{h}_{\mathcal{S}}(x)$ & Compact forensic subspace formed by concatenating the activations of selected LANs. \\
$\mathbf{z}(x)$; $P_{\mathcal{A}}$ & Concatenated activations in the localized candidate space and projection onto a neuron subset $\mathcal{A}$. \\
$f_{\mathcal{S}}$; $\mathbf{w}_{\mathcal{S}}$; $b_{\mathcal{S}}$ & Linear readout on the selected LAN subspace and its parameters. \\
$\boldsymbol{\mu}_{\mathrm{real}}$; $\boldsymbol{\mu}_{\mathrm{fake}}$; $\Delta\boldsymbol{\mu}$; $\Sigma$ & Class centroids, their difference, and the pooled covariance in the localized candidate space. \\
$\boldsymbol{\delta}^{(\ell)}$ & Layer-wise real--fake centroid displacement, defined as $\boldsymbol{\mu}_{\mathrm{fake}}^{(\ell)}-\boldsymbol{\mu}_{\mathrm{real}}^{(\ell)}$. \\
$J(\mathcal{A})$ & Fisher-style separability score of coordinate subset $\mathcal{A}$. \\
$J_\ell$ & Fisher-style separability score of layer $\ell$ under the locally isotropic approximation. \\
$\bar{\mathcal{S}}$; $\eta$ & Complement of the selected LAN set and the residual mean-gap energy ratio bound. \\
$f^{\star}$; $\mathbf{w}^{\star}$; $b^{\star}$ & Reference linear separator and its parameters. \\
$\mathbf{w}^{\star}_{\mathcal{S}}$; $\mathbf{w}^{\star}_{\bar{\mathcal{S}}}$ & Reference-separator weights restricted to selected and discarded coordinates, respectively. \\
$\mathrm{mACC}$; $\mathrm{mAP}$; $\mathrm{AUC}$ & Mean accuracy, mean average precision, and area under the ROC curve. \\
\hline
\end{tabular}
\end{table*}

\section{Experimental Protocol Details}
\label{ap1}
\noindent \textbf{Role of the Internal Splits.}
This section only supplements protocol details that are not fully expanded in the main paper. From the 15Model-140K source detector training distribution~\citep{li2026preserving}, we construct three mutually disjoint internal splits containing 2400, 2400, and 1200 videos, respectively. The discovery split is used exclusively for layer localization and LAN scoring, the training split is used exclusively to fit the final linear classifier after neuron selection, and the validation split is used only for model selection within the source distribution. No external benchmark is involved in layer localization, neuron discovery, or checkpoint selection in our protocol.

\noindent \textbf{Validation Usage and Checkpoint Selection.}
The validation split is not treated as an additional reporting benchmark, but only as a model-selection set for internal choices such as checkpoint ranking. External evaluation is performed on Magic Videos~\citep{li2026preserving}, MovieGen~\citep{polyak2024moviegen}, and DVF~\citep{song2024multimodal}. In addition, all experiments in the main Qwen2.5-ViT setting use the same released detector checkpoint~\citep{li2026preserving} and keep the backbone frozen, so the comparison focuses on whether a compact LAN subspace can preserve forgery-discriminative knowledge without additional backbone adaptation.

\noindent \textbf{Fairness Constraints in Fixed-Budget Controls.}
The main paper already reports the common backbone, preprocessing settings, and optimization hyperparameters, so we only clarify here the fairness constraints used in the control experiments. All fixed-budget comparisons use the same feature budget as the main setting. In particular, the same-layer random baseline preserves the layer-wise allocation of the selected LAN set, which prevents the comparison from being affected by a different layer budget or a larger effective search space. This design makes the control experiments test neuron informativeness rather than feature dimensionality alone.

\section{Deferred Mathematical Definitions}
\label{ap2}
This section provides the exact statistical definitions that are only briefly referenced in the main paper for readability. We specifically complete the formulas used in layer localization and latent anchor neuron discovery.

\noindent \textbf{Definitions for Layer Localization.}
The normalized-shift metric $D_{\mathrm{shift}}(\ell)$ in the main paper depends on the mean within-class channel variance at layer $\ell$ over all latent channels, which is defined as
\begin{equation}
\bar{\sigma}_\ell^2=\frac{1}{2M}\sum_{c\in\{\mathrm{real},\mathrm{fake}\}}\sum_{k=1}^{M}\mathrm{Var}_{x\in\mathcal{D}_c}\!\big[a_k^{(\ell)}(x)\big].
\end{equation}
This quantity averages channel-wise within-class variation over both classes and all $M$ latent units, and is used to normalize centroid displacement across layers with different activation scales. The two layer-localization thresholds are computed from the layer-wise score distributions on the discovery split:
\begin{equation}
\tau_{\mathrm{sep}} = \mathrm{mean}\!\left(D_{\cos}\right)+\mathrm{std}\!\left(D_{\cos}\right).
\end{equation}
\begin{equation}
\tau_{\mathrm{shift}} = \mathrm{mean}\!\left(D_{\mathrm{shift}}\right)+\mathrm{std}\!\left(D_{\mathrm{shift}}\right).
\end{equation}
The critical-layer set is then obtained by intersecting the two thresholded sets.

\noindent \textbf{Definitions for LAN Discovery.}
The neuron separation effect size $d_n$ uses the pooled standard deviation of the real and fake response groups. For class $c\in\{\mathrm{real},\mathrm{fake}\}$, we define the weight-normalized within-class response variance of neuron $n$ as
\begin{equation}
\sigma_{n,c}^{2}
=
\frac{1}{|\mathcal{D}_c|-1}
\sum_{x\in\mathcal{D}_c}
\left(
\frac{r_n(x)-\mu_{n,c}}{|w_n|+\epsilon}
\right)^2.
\end{equation}
The pooled standard deviation is then given by
\begin{equation}
\sigma_n^{\mathrm{pool}}=
\sqrt{
\frac{
\left(|\mathcal{D}_{\mathrm{real}}|-1\right)\sigma_{n,\mathrm{real}}^2+
\left(|\mathcal{D}_{\mathrm{fake}}|-1\right)\sigma_{n,\mathrm{fake}}^2
}{
|\mathcal{D}_{\mathrm{real}}|+|\mathcal{D}_{\mathrm{fake}}|-2
}
}.
\end{equation}

\section{Theoretical Analysis of V-FIND}
\label{ap3}
We provide a theoretical interpretation of V-FIND in a form aligned with the actual pipeline used in the paper. The video-forgery backbone is kept frozen, layer localization restricts the candidate space, LAN discovery selects a sparse subset of activations, and only a lightweight linear readout is optimized. The analysis shows that layer localization favors layers with large dimension-normalized class separability, whereas sparse LAN selection preserves the dominant real--fake signal when the mean-gap energy is concentrated on the selected coordinates.

\noindent \textbf{Problem Formulation.}
Let
\begin{equation}
\mathbf{z}(x)=[a_n(x)]_{n\in\mathcal{N}^{\star}}
\end{equation}
denote the concatenated activations from the localized candidate set, and let $P_{\mathcal{A}}$ denote the coordinate projection onto a neuron subset $\mathcal{A}$. For the selected LAN set $\mathcal{S}$, the final feature and readout are
\begin{equation}
\mathbf{h}_{\mathcal{S}}(x)=P_{\mathcal{S}}\mathbf{z}(x),
\qquad
f_{\mathcal{S}}(x)=\mathbf{w}_{\mathcal{S}}^\top \mathbf{h}_{\mathcal{S}}(x)+b_{\mathcal{S}}.
\end{equation}
We denote the class means in the localized candidate space by $\boldsymbol{\mu}_{\mathrm{real}}$ and $\boldsymbol{\mu}_{\mathrm{fake}}$, define $\Delta\boldsymbol{\mu}=\boldsymbol{\mu}_{\mathrm{fake}}-\boldsymbol{\mu}_{\mathrm{real}}$, and let $\Sigma$ be the pooled covariance. For any coordinate subset $\mathcal{A}$, a Fisher-style separability score can be written as
\begin{equation}
J(\mathcal{A})
=
{\left(P_{\mathcal{A}}\Delta\boldsymbol{\mu}\right)}^\top
{\left(P_{\mathcal{A}}\Sigma P_{\mathcal{A}}^\top\right)}^{-1}
\left(P_{\mathcal{A}}\Delta\boldsymbol{\mu}\right).
\end{equation}
The objective is to characterize when $J(\mathcal{S})$ remains large after projecting the localized representation onto a much smaller LAN subspace.

\noindent \textbf{Layer Localization.}
For layer $\ell$, let $\boldsymbol{\delta}^{(\ell)}=\boldsymbol{\mu}_{\mathrm{fake}}^{(\ell)}-\boldsymbol{\mu}_{\mathrm{real}}^{(\ell)}$. Under a locally isotropic approximation $\Sigma^{(\ell)}\approx\bar{\sigma}_\ell^2 I$, the dimension-normalized layer separability satisfies
\begin{equation}
\begin{aligned}
\frac{1}{M}J_\ell
&=
\frac{1}{M}
{\boldsymbol{\delta}^{(\ell)}}^\top
{\Sigma^{(\ell)}}^{-1}
\boldsymbol{\delta}^{(\ell)}
\\
&\approx
\frac{\|\boldsymbol{\delta}^{(\ell)}\|_2^2}{M\bar{\sigma}_\ell^2}
=
D_{\mathrm{shift}}(\ell)^2.
\end{aligned}
\end{equation}
Thus, $D_{\mathrm{shift}}(\ell)$ estimates the real--fake displacement after accounting for within-class fluctuation and layer width. The complementary score $D_{\cos}(\ell)$ further measures whether the two class centroids diverge in direction. The intersection
\begin{equation}
\mathcal{L}^{\star}=\mathcal{L}_{\mathrm{sep}}\cap\mathcal{L}_{\mathrm{shift}}
\end{equation}
therefore acts as a conservative filter in the layer-selection stage that retains layers with both directional separation and normalized class displacement.

\noindent \textbf{Sparse LAN Readout.}
Within the localized candidate layers, V-FIND retains neurons with large response effect size
\begin{equation}
d_n=\frac{|\mu_{n,\mathrm{fake}}-\mu_{n,\mathrm{real}}|}{\sigma_n^{\mathrm{pool}}+\epsilon},
\end{equation}
which favors coordinates whose class gap is strong relative to their weight-normalized within-group variation. Let $\bar{\mathcal{S}}$ denote the complement of $\mathcal{S}$ within the localized candidate set. If the discarded coordinates contain only a bounded amount of mean-gap energy,
\begin{equation}
\left\|P_{\bar{\mathcal{S}}}\Delta\boldsymbol{\mu}\right\|_2
\leq
\eta
\left\|P_{\mathcal{S}}\Delta\boldsymbol{\mu}\right\|_2,
\qquad \eta < 1,
\end{equation}
then the selected coordinates preserve a lower-bounded fraction of the total mean-gap energy:
\begin{equation}
\frac{\left\|P_{\mathcal{S}}\Delta\boldsymbol{\mu}\right\|_2^2}
{\left\|\Delta\boldsymbol{\mu}\right\|_2^2}
\geq
\frac{1}{1+\eta^2}.
\end{equation}
This condition does not require all localized channels to be retained. It only requires that the main class-discriminative displacement be concentrated on the selected LAN coordinates, which is precisely what the effect-size criterion is designed to approximate.

Consider a reference linear separator on the localized representation,
$f^{\star}(\mathbf{z})={\mathbf{w}^{\star}}^\top\mathbf{z}+b^{\star}$.
The discrepancy between the full localized separator and its restriction to the selected LAN coordinates is bounded by
\begin{equation}
\left|f^{\star}(\mathbf{z}(x))-\left({\mathbf{w}^{\star}_{\mathcal{S}}}^\top \mathbf{h}_{\mathcal{S}}(x)+b^{\star}\right)\right|
\leq
\left\|\mathbf{w}^{\star}_{\bar{\mathcal{S}}}\right\|_2
\left\|\mathbf{z}_{\bar{\mathcal{S}}}(x)\right\|_2.
\end{equation}
Therefore, if the omitted-coordinate contribution is smaller than the classification margin,
\begin{equation}
\left\|\mathbf{w}^{\star}_{\bar{\mathcal{S}}}\right\|_2
\left\|\mathbf{z}_{\bar{\mathcal{S}}}(x)\right\|_2
<
\left|f^{\star}(\mathbf{z}(x))\right|,
\end{equation}
the sparse projection preserves the sign of the reference decision. This gives a margin-based condition under which a frozen backbone with a lightweight readout remains effective after projection onto the selected LANs.

This analysis is consistent with the main observations in the paper and the appendix. Layer localization reduces the search space to a small set of discriminative layers, selected LANs outperform same-budget random neurons from the same layers, and the lightweight sparse readout maintains strong mACC and mAP with only a compact subset of activations. Together, these observations support the interpretation that V-FIND exposes a compact forensic subspace already encoded in the frozen detector. We regard this analysis as an explanatory approximation rather than a formal optimality proof for the nonlinear detector.

\section{More Experiments}
\label{ap4}
\begin{figure}[t]
    \centering
    \includegraphics[width=\columnwidth]{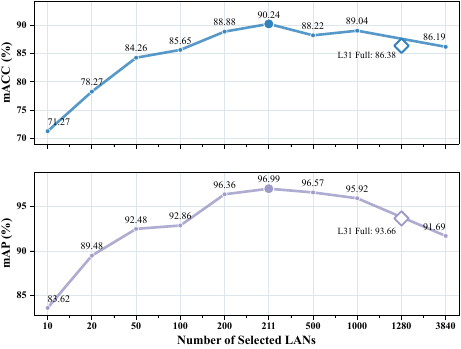}
    \vspace{-0.3cm}
    \caption{\textbf{Sensitivity to the Number of Selected LANs.} Performance improves substantially from very small subsets to moderate-size subsets, reaches its strongest region near the main setting, and drops again when all localized channels are used. The open diamond at 1280 denotes the layer-31 full-head control, which uses all 1280 channels of the final layer.}
    \label{fig:appendix-lan-count-sensitivity}
\end{figure}

\noindent \textbf{Sensitivity to the Number of Selected LANs.}
We first vary the number of selected LANs while keeping the rest of the evaluation protocol unchanged. Figure~\ref{fig:appendix-lan-count-sensitivity} shows that performance improves rapidly from very small subsets to moderate-size subsets, reaches its strongest region near the main setting, and then becomes comparatively stable. By contrast, replacing the sparse readout with all localized channels gives only 86.19 mACC and 91.69 mAP on Magic. A full linear head on all 1280 channels of layer 31 gives 86.38 mACC and 93.66 mAP. Both controls remain below the main sparse readout (90.24 mACC and 96.99 mAP), supporting the view that V-FIND benefits from retaining a compact discriminative subset rather than simply enlarging the feature dimension.

\begin{figure}[t]
    \centering
    \includegraphics[width=\columnwidth]{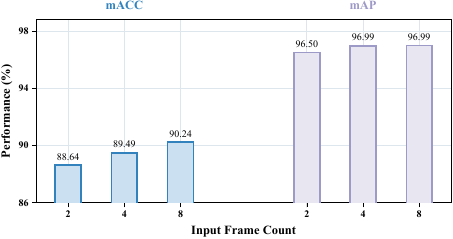}
    \vspace{-0.3cm}
    \caption{\textbf{Sensitivity to the Input Frame Count.} Bars are grouped by metric: the left group compares mACC across 2, 4, and 8 frames, and the right group compares mAP across the same frame counts. The 4-frame and 8-frame settings remain very close.}
    \label{fig:appendix-frame-count-sensitivity}
\end{figure}

\noindent \textbf{Sensitivity to the Input Frame Count.}
We further test whether the readout is sensitive to the number of input frames used at inference time. In this analysis, only the sampled frame count is varied among 2, 4, and 8. Figure~\ref{fig:appendix-frame-count-sensitivity} reports the two metrics as separate groups, so that mACC is compared only across frame counts and mAP is compared in the same way. On Magic, the readout obtains 88.64 mACC and 96.50 mAP with 2 frames, 89.49 and 96.99 with 4 frames, and 90.24 and 96.99 with 8 frames. The 4-frame and 8-frame results remain very close, indicating that the observed performance is not tied to a single narrowly tuned frame count.

\section{Further Discussions}
\label{ap5}
\noindent \textbf{Q1: \textit{Why does V-FIND keep the detector backbone frozen?}}
The objective of V-FIND is to examine whether a released video forgery detector already contains extractable forgery-discriminative knowledge, rather than to obtain another detector through additional end-to-end optimization. Updating the backbone would mix two effects: discovering existing internal knowledge and learning new representations from the source distribution. We therefore keep the original detector fixed during layer localization, neuron discovery, and final readout training. Under this setting, the performance of the lightweight classifier can be more directly attributed to the localized forensic subspace. This design also makes the intervention analysis more meaningful, because changes caused by manipulating selected LANs reflect the functional role of neurons already encoded in the frozen detector.

\noindent \textbf{Q2: \textit{Why are both layer localization and LAN discovery necessary?}}
The two stages address different granularities in our pipeline and are not redundant. Layer localization first asks where real--fake discrepancy is concentrated across the backbone, using complementary directional and shift-based signals to avoid searching all layers uniformly. However, a localized layer is still a high-dimensional representation that contains many units unrelated to the final forgery decision. LAN discovery therefore further asks which neurons within the localized layers consistently carry discriminative responses. This distinction is supported by the fixed-budget controls and the appendix sensitivity analysis: same-layer random features improve over global random features, but the selected LAN subspace remains stronger; using all localized channels also does not simply improve the result. Thus, V-FIND relies on both coarse layer filtering and fine neuron selection to isolate a compact and functional forensic subspace within the frozen detector.

\noindent \textbf{Q3: \textit{Is V-FIND a detector compression method or a forensic-subspace analysis method?}}
V-FIND is closer to a forensic-subspace analysis method than to a detector compression method. Although the final readout uses only a compact set of LANs, the purpose is not to replace the original detector with the smallest possible model. Instead, the sparse readout serves as a controlled probe for testing whether a compact set of selected neurons encodes more forgery-discriminative knowledge that is extractable and supports an effective task-relevant readout from the frozen backbone. This distinction is important because compression would primarily be evaluated by efficiency, latency, or parameter reduction, whereas V-FIND is evaluated by whether the selected neurons remain discriminative, outperform matched random controls, and produce meaningful intervention effects. Therefore, the compactness of the LAN readout should be interpreted as evidence about the organization of forensic signals inside the detector, rather than as a claim that the full detector can be universally compressed without loss.

\section{Limitations and Future Directions}
\label{ap6}
Although our experiments cover multiple generator sources and complementary control settings, we do not conduct a systematic investigation of closely related detector variants within the same model family. Consequently, whether the localized LAN distributions and their relative layer-wise concentration remain fully consistent across detector checkpoints, training scales, and backbone variants has yet to be established. Future work will further examine the intra-family consistency and transferability of LANs across related video forgery detectors.

\end{document}